# ADBSCAN: Adaptive Density-Based Spatial Clustering of Applications with Noise for Identifying Clusters with Varying Densities

Mohammad Mahmudur Rahman Khan[1*], Md. Abu Bakr Siddique[2#], Rezoana Bente Arif[2@], and Mahjabin Rahman Oishe[3$]
[1]Dept. of ECE, Mississippi State University, Mississippi State, MS 39762, USA
[2]Dept. of EEE, International University of Business Agriculture and Technology, Dhaka 1230, Bangladesh
[3]Dept. of CSE, Rajshahi University of Engineering and Technology, Rajshahi 6204, Bangladesh
mrk303@msstate.edu[*], absiddique@iubat.edu[#], rezoana@iubat.edu[@], mahjabinoishe@gmail.com[$]

*Abstract*—Density-based spatial clustering of applications with noise (DBSCAN) is a data clustering algorithm which has the high-performance rate for dataset where clusters have the constant density of data points. One of the significant attributes of this algorithm is noise cancellation. However, DBSCAN demonstrates reduced performances for clusters with different densities. Therefore, in this paper, an adaptive DBSCAN is proposed which can work significantly well for identifying clusters with varying densities.

## I. INTRODUCTION

In computer science and machine learning, data mining [1] is the criteria for extracting useful and accurate information from large datasets combining tools from statistics, artificial intelligence and database systems [2]. It is a sub-discipline of computer science where fields from other science merge [3]. It takes advantages of inventive methods to extract hidden data patterns and establish relationships to solve the problem through data analysis [2, 4, 5]. A data mining problem mainly comprises of six everyday tasks [1] including anomaly detection, dependency modeling, clustering, classification, regression, and summarization.

The branch of data mining called clustering [6-8] is very handy in extracting usable and accurate information from data. Clustering is the process of recognizing and making a group similar objects into subsets to locate homogeneous groups of objects depending upon the values of their features and attributes [9]. It is one of the most useful unsupervised learning method having many applications in various domains inclusive of biology, medicine, medical imaging [10], business and marketing, image segmentation [11], robotics [12], analytical chemistry [13], climatology [14], etc. Though there are several published clustering algorithms in recent years, the most prominent and used clustering algorithms can be widely categorized as hierarchical [15-17], partitioning [18-20] and density based [21-24] methods.

Density-based clustering is the method of identifying distinctive groups or clusters in a dataset relied on the notion that a cluster is a dense contiguous region in the total data space, which separated from other clusters by adjacent areas of relatively lower data density [25]. The data points having a comparatively lower object density in the separating regions are typically labeled as noise or outliers [25].

DBSCAN [21] is deemed as one of the most powerful and most cited density-based clustering algorithms which can identify with significant accuracy the clusters of random shape and size in large databases corrupted with noise. One of the main advantages of the DBSCAN algorithm is that predetermination of the number of clusters is not required on datasets [26]. As the DBSCAN algorithm is capable of handling the noise points correctly and effectively [27], it is more applicable to find a group surrounded by noise as well as different other groups [26]. Again, it is regarded as insensitive to noise and outliers. However, two critical parameters are the basic requirement for applying the DBSCAN algorithm: i) Eps and ii) MinPts. Eps is the radius of the adjacent neighborhood of a considered data point, and MinPts is the adjacent minimum number of data points located in the given region. Though users do not need adequate pre-knowledge of the number of clusters in the datasets, Eps and MinPts are determined by the users [28]. As dataset differs from each other in a large range, it is difficult to predict an appropriate Eps value manually. But as Eps plays a significant role in the outputs of the DBSCAN algorithm, the results may vary dramatically for different values of Eps. As a result, the overall performance of DBSCAN degrades considerably in the case of finding the clusters with differing data densities, because the Eps value, in DBSCAN, is a global predetermined parameter [29]. Therefore, for identifying the cluster with varying densities, separate Eps, and corresponding MinPts value should be adapted according to the data distribution.

In this paper, an adaptive DBSCAN (ADBSCAN) algorithm is proposed to determine an appropriate Eps and MinPts values so that the algorithm can identify all the clusters in the datasets. This algorithm first starts with a random value of Eps. If it fails to detect a cluster, it increases the Eps value 0.5 and so on. If in an iteration, over 10% of similar data has been identified, the algorithm assumes that a cluster has been discovered. Then that cluster is saved separately and excluded from the primary dataset. The algorithm again increases the Eps as well as MinPts values to detect the next group and so on. In this way, when above 95% data will be covered; it is considered that all the clusters have been identified successfully. Then the remaining data points will be declared as noise points or outliers, and the saved clusters will be plotted.

## II. DBSCAN Algorithm's Overview and Implementation

DBSCAN is a density-based clustering algorithm. Let the dataset be denoted as D, the algorithm's clustering radius, Eps, and the minimum number of objects in the Eps-neighborhood, MinPts, and then the basic concepts of the algorithm can be expressed by the following definitions [21]:

<u>Eps-neighborhood:</u> The neighboring data point within the radius of Eps of a given point in the dataset is called the Eps-neighborhood of that point.

<u>Core point:</u> If the Eps-neighborhood of a test point contains a minimum number of neighbors, MinPts, then the test point is considered a core point.

<u>Border point:</u> If the Eps-neighborhood of a test point does not contain a minimum number of neighbors, MinPts, but it is identified to share its Eps-neighborhood with at least one identified core point, then the test point is called a border point.

<u>Directly density reachable:</u> A point p is deemed as directly density-reachable from another point q if p is within Eps-neighborhood of q and q is identified as a core point. The definition of directly density reachability is demonstrated in figure 1.

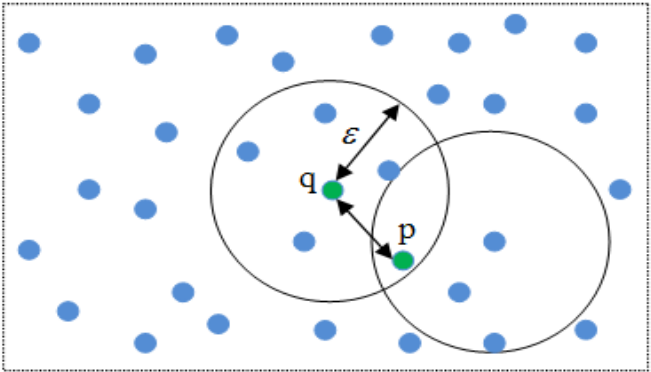

Fig. 1. Point p is directly density-reachable from point q

<u>Density reachable:</u> A point p is deemed as density-reachable from another point q according to Eps and MinPts in a dataset, D, if there happens to be a chain of points $p_1, p_2, \ldots, p_n$, $p_1 = q$ and $p_n = p$ in a way that $p_{i+1}$ is directly density-reachable from $p_i$, regarding Eps and MinPts, for $1 \leq i \leq n, p_i \in D$. The definition of density reachable is demonstrated in figure 2.

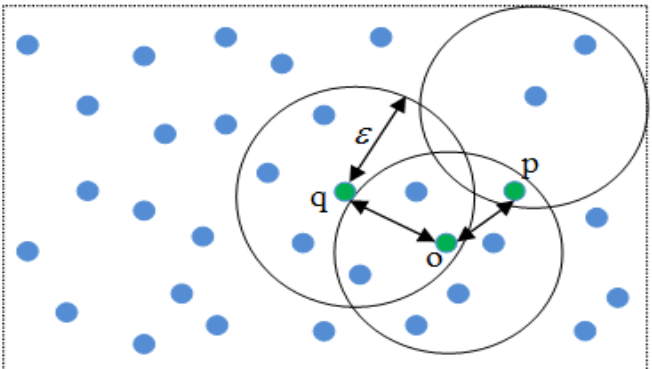

Fig. 2. Point p is density-reachable from point q

<u>Density connected:</u> A point p is deemed as density connected to another point q according to Eps and MinPts in a dataset, D if there exist a data point $o \in D$ such that both p and q are density-reachable from point o regarding Eps and MinPts. The definition of density connected is demonstrated in figure 3.

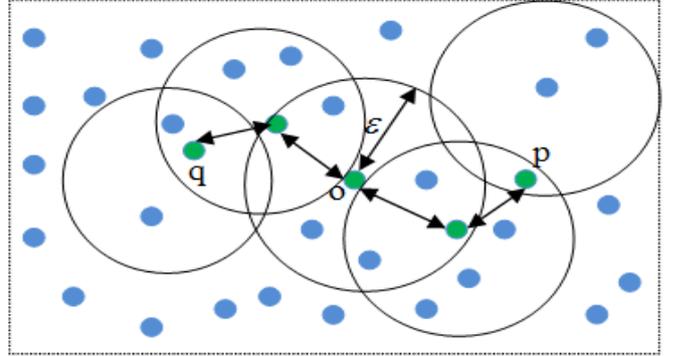

Fig. 3. Point p is density connected to point q

DBSCAN searches for clusters in a dataset by checking Eps-neighborhood of each test point in the dataset. If the Eps-neighborhood of a test data point contains more than a minimum number of neighbors, MinPts, a new cluster with point p as a core is created. The algorithm then assembles directly density reachable points from these core points to merge all the density reachable clusters. This process terminates when there is no new point to be added to any group. The points which do not fall in any cluster are considered to be noise or outliers. The following is the algorithm of the DBSCAN:

- DBSCAN begins with a random point $p$
- It retrieves all points that are density-reachable from p for distance, Eps $(\varepsilon)$, which are known as MinPts.
- If p is identified as a core point, this method considers a cluster w.r.t. Eps $(\varepsilon)$ and MinPts
- If p is identified as a border point where no points are found to be density-reachable from p, and DBSCAN moves to the next data point of the database.

The working methodology of DBSCAN is briefly presented in the following flowchart as demonstrated in figure 4.

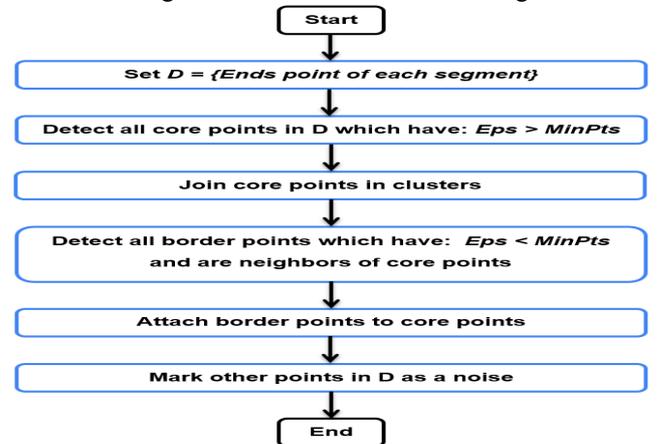

Fig. 4. The flowchart of the basic DBSCAN algorithm

At this first point of the algorithm development, we have generated a small synthetic dataset with two clusters for testing purpose. The dataset is given in figure 5.

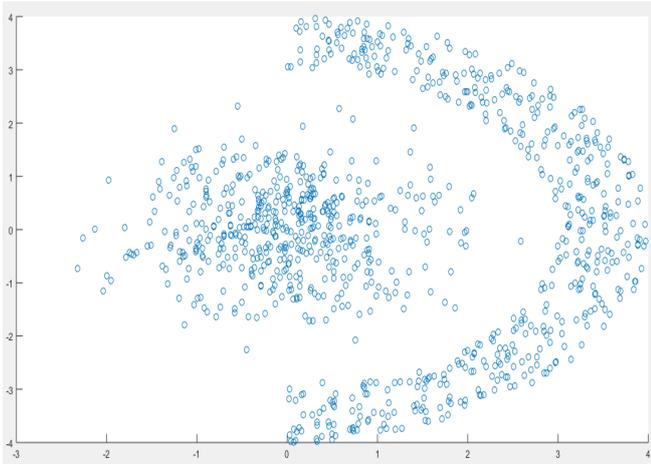

Fig. 5. The dataset with two clusters of similar data density

Here, it is visible that the data set contains two clusters. We implemented the DBSCAN clustering algorithm to observe the performance. The two clusters have almost similar distribution densities. Two important parameters of DBSCAN are $\varepsilon$ and MinPts. The $\varepsilon$ means the radius of the neighborhood for point P and MinPts denotes the minimum number of data points in the given neighborhood. Now, while implementing DBSCAN, these parameters are considered to have constant values which limit the DBSCAN algorithm to perform for datasets with varying densities. In this case, we have selected, Eps = 0.5 and MinPts = 10. The DBSCAN algorithm identified two clusters and some noises in the dataset. The performance of the DBSCAN algorithm is depicted in figure 6.

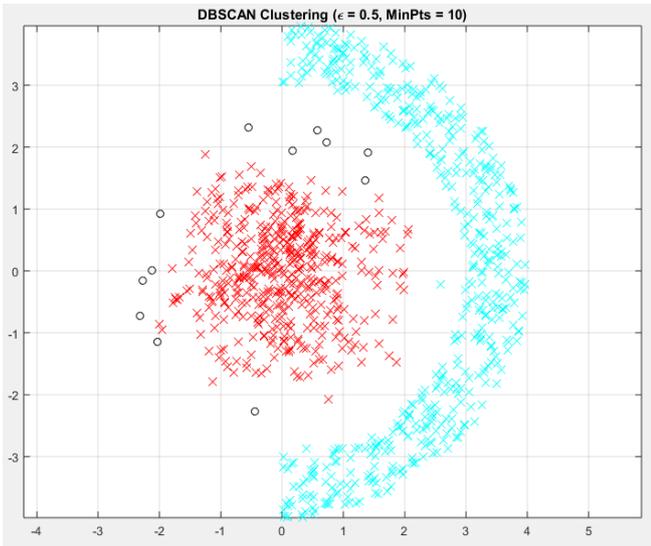

Fig. 6. The performance of DBSCAN with two clusters of similar data density

Now, as the clusters had similar distribution densities, the DBSCAN was very useful here. However, when we tried to implement clusters with varying distribution densities, then the algorithm started giving inferior performances as presented in figure 7 and 8 for three and four clusters of varying data density respectively. For both cases, we have taken Eps = 5 and MinPts = 10. For both three and four clusters datasets, DBSCAN only recognized one cluster and remaining data identified as noise.

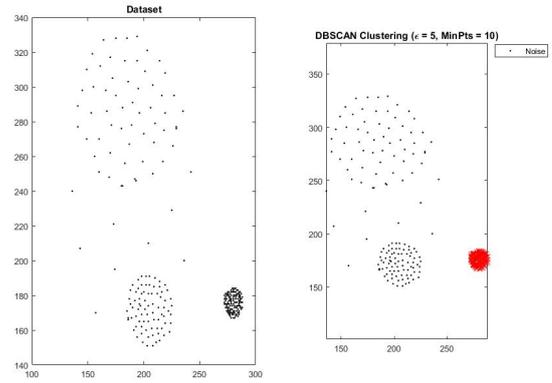

Fig. 7. The performance of DBSCAN with three clusters of varying data density

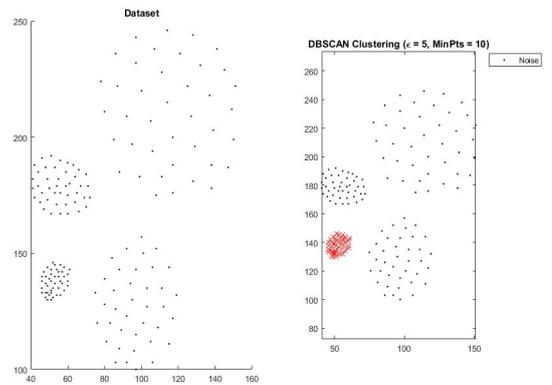

Fig. 8. The performance of DBSCAN with four separate clusters of varying data density

### III. PROPOSED APPROACH: ADBSCAN FOR CLUSTERING

To overcome the difficulties of DBSCAN algorithm to discover clusters in a dataset with varying data densities, in the next steps, we modified the existing DBSCAN algorithm so that it can adapt the values of Eps and MinPts on the basis of the density distribution of the clusters. Adaptive DBSCAN (ADBSCAN) algorithm can determine an appropriate Eps and MinPts value automatically. The ADBSCAN algorithm first starts with a random value of Eps. If it fails to detect a cluster, it increases the Eps value 0.5. If in an iteration, over 10% of similar data has been identified, it is considered that a cluster has been discovered. Then that cluster has been saved separately and excluded from the primary dataset. The algorithm again increases the Eps as well as MinPts values to detect the next cluster. In this way, when 95% data has been exhausted, the algorithm assumes that all the clusters have been detected successfully. Then the remaining data points will be declared as noise points or outliers, and the saved clusters will be plotted. For making the DBSCAN algorithm adaptive to the density variation among various clusters, we have modified the basic DBSCAN algorithm rendering to the following flowchart as given in figure 9.

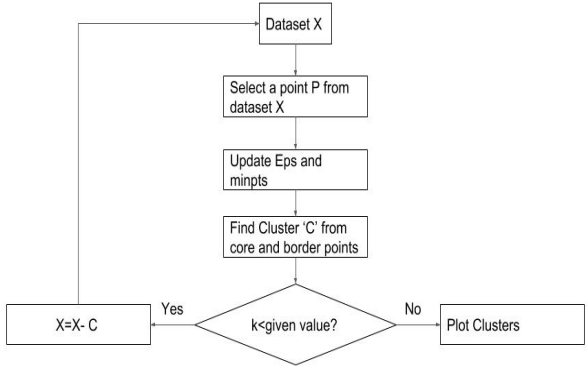

Fig. 9. The flowchart of the adaptive DBSCAN algorithm

Unlike DBSCAN, ADBSCAN requires predetermination of the number of data clusters in a dataset. Here, $k=1,2,...,C$. $C$ = the number of data clusters in the datasets. In regards to adaptive DBSCAN, we start with the dataset X and initial values of Eps, $\varepsilon$ and Minpts. At first, we consider an arbitrary point P. Then we update our $\varepsilon$ and Minpts to get the densest cluster C. Once we get the densest cluster, we save the data points in a group and remove those points from the dataset. Then we update $\varepsilon$ and Minpts again to get the next dense cluster. That is how we get all the clusters which have varying densities one by one. The algorithm of adaptive DBSCAN is given below in figure 10.

**Algorithm 1 ADBSCAN Algorithm**
**Input:** Dataset $X$, Total number of clusters: $K$
**Assume:** Eps and minpts
Select a point, $P$
**Update:** Eps and minpts
$Loop = 1$
**if** Recognized Points $> 10\%$ **then**
   Identify core and border points for cluster, $C$
   $Loop++$
   $X = X - C$
   **while** $Loop < K$ **do**
     Repeat the process to find the next cluster
   **end while**
**else**
   Increase Eps and minpts by 0.5 and repeat
   **if** $X \leq 5\%$ **then**
     Plot all the clusters
   **end if**
**end if**

Fig. 10. Adaptive DBSCAN clustering algorithm

## IV. RESULTS AND DISCUSSION

Since this is going to be a new method; therefore, in the initial stage, we want to study the performance of adaptive DBSCAN algorithm by using the state-of-the-art synthetic datasets, which are widely utilized in different kinds of literature. If the trial clustering process on these synthetic datasets is successful, then, we can extend our analysis to the more massive datasets that are used in cloud computing purposes.

The proposed method was applied to the following dataset with three clusters of varying data density as presented in figure 11. It is visible that the dataset has three clusters and some noises. Our adaptive DBSCAN showed excellent performance in case of identifying all the clusters and the noise, whereas, the conventional DBSCAN is only able to find one cluster.

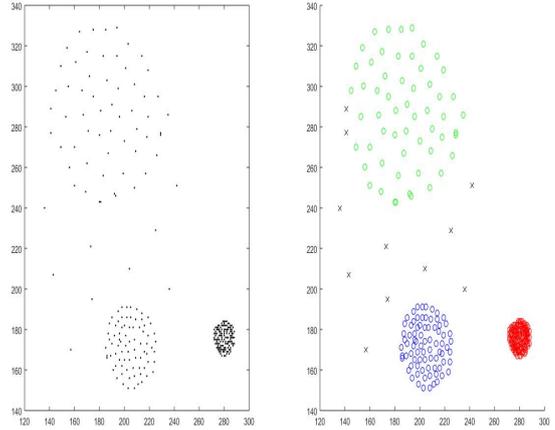

Fig. 11. The performance of ADBSCAN with three clusters of varying data density

Then we applied a dataset with four separate clusters of varying data density as demonstrated in figure 12. The proposed adaptive DBSCAN shows promising performance here too, whereas, the conventional DBSCAN is only able to find one cluster.

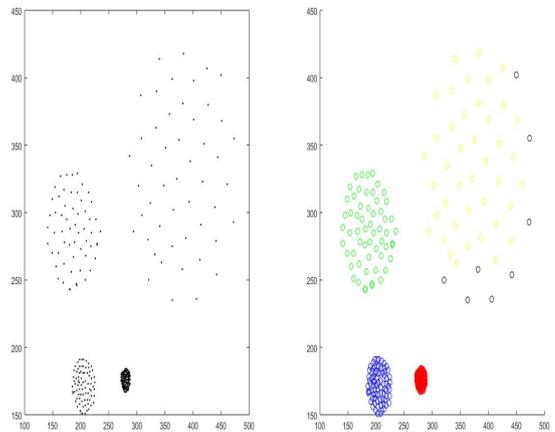

Fig. 12. The performance of ADBSCAN with four separate clusters of varying data density

After careful observations from the results produced by the proposed ABDSCAN method as demonstrated in figures as mentioned above, it is clear that the ADBSCAN performs with better accuracy compared to basic DBSCAN in case of cluster recognition where the clusters have different data densities. ADBSCAN demonstrates reasonably better performance in case of noisy data as well.

## V. CONCLUSION

From the results and discussion section, it is evident that the proposed adaptive DBSCAN outperforms conventional

DBSCAN in case of clusters with varying densities. However, the proposed clustering method has some limitations. The method initializes with random values of Eps, $\varepsilon$ and Minpts which makes it less generalized. Our next goal is to investigate and improve the adaptive DBSCAN algorithm in a way that it would be able to pinpoint the optimal values of $\varepsilon$ and Minpts for each cluster.